\documentclass{article}

\usepackage[final]{corl_2020} 

\usepackage{multicol}

\usepackage{epsfig}
\usepackage{amsthm}
\usepackage{epstopdf}
\usepackage{graphicx}
\graphicspath{{Figures/}}
\usepackage{url}
\usepackage{stfloats}
\usepackage{amsmath}
\usepackage{amsfonts}   
\usepackage{amsopn}     
\usepackage{psfrag}
\usepackage[tight]{subfigure}
\usepackage{mdwtab}     
\usepackage{enumerate}
\usepackage{color}
\usepackage{accents}
\usepackage{mathtools}
\usepackage{algorithm}
\usepackage{algpseudocode}
\usepackage{adjustbox}
\usepackage{cases}

\newtheorem{theorem}{\bf Theorem}
\newtheorem{prop}{\bf Proposition}

\newtheorem{assumption}{\bf Assumption}

\usepackage{wrapfig}

\renewcommand{\baselinestretch}{0.99}
\usepackage{titlesec}
\usepackage{etoolbox}

\makeatletter
\patchcmd{\ttlh@hang}{\parindent\z@}{\parindent\z@\leavevmode}{}{}
\patchcmd{\ttlh@hang}{\noindent}{}{}{}
\makeatother
\titlespacing*{\section}{0pt}{0.2ex}{0.2ex}
\titlespacing*{\subsection}{0pt}{0.2ex}{0.2ex}
\titlespacing*{\subsubsection}{0pt}{0.2ex}{0.2ex}

\title{\LARGE \bf
Probably Approximately Correct Vision-Based Planning using Motion Primitives
}

\author{
  Sushant Veer\\
  Mechanical and Aerospace Engineering\\
  Princeton University\\
  \texttt{sveer@princeton.edu}
  \And
  Anirudha Majumdar \\
  Mechanical and Aerospace Engineering\\
  Princeton University\\
  \texttt{ani.majumdar@princeton.edu}
}

\input{irom.sty}

\begin{document}
\setlength{\textfloatsep}{3mm}
\setlength{\abovedisplayskip}{2pt}
\setlength{\belowdisplayskip}{2pt}
\setlength{\abovedisplayshortskip}{1pt}
\setlength{\belowdisplayshortskip}{1pt}

\maketitle

\begin{abstract}
This paper presents an approach for learning vision-based planners that provably generalize to novel environments (i.e., environments unseen during training). We leverage the \emph{Probably Approximately Correct (PAC)-Bayes} framework to obtain an upper bound on the expected cost of policies across all environments. Minimizing the PAC-Bayes upper bound thus trains policies that are accompanied by a \emph{certificate of performance} on novel environments. The training pipeline we propose provides strong generalization guarantees for deep neural network policies by (a) obtaining a good prior distribution on the space of policies using Evolutionary Strategies (ES) followed by (b) formulating the PAC-Bayes optimization as an efficiently-solvable parametric \emph{convex optimization} problem. We demonstrate the efficacy of our approach for producing strong generalization guarantees for learned vision-based motion planners through two simulated examples: (1) an Unmanned Aerial Vehicle (UAV) navigating obstacle fields with an onboard vision sensor, and (2) a dynamic quadrupedal robot traversing rough terrains with proprioceptive and exteroceptive sensors. \end{abstract}


\section{Introduction}
\label{sec:intro}

\begin{figure*}[t]
\vspace{-2mm}
\centering
\subfigure[]
{
\includegraphics[width=0.43\textwidth]{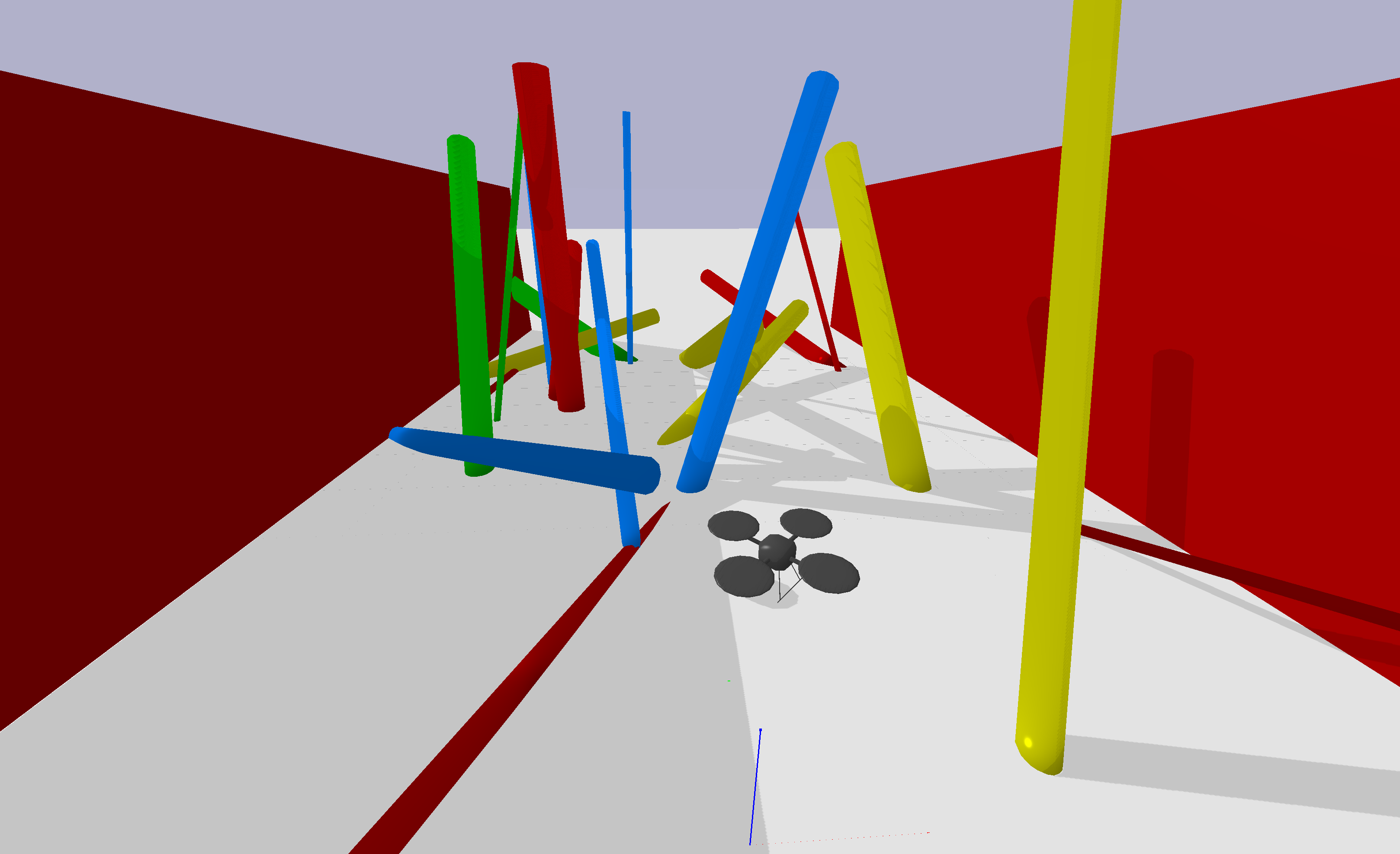}
\label{fig:quadrotor}
}
\centering
\hspace{2mm}
\subfigure[]
{
\includegraphics[width=0.43\textwidth]{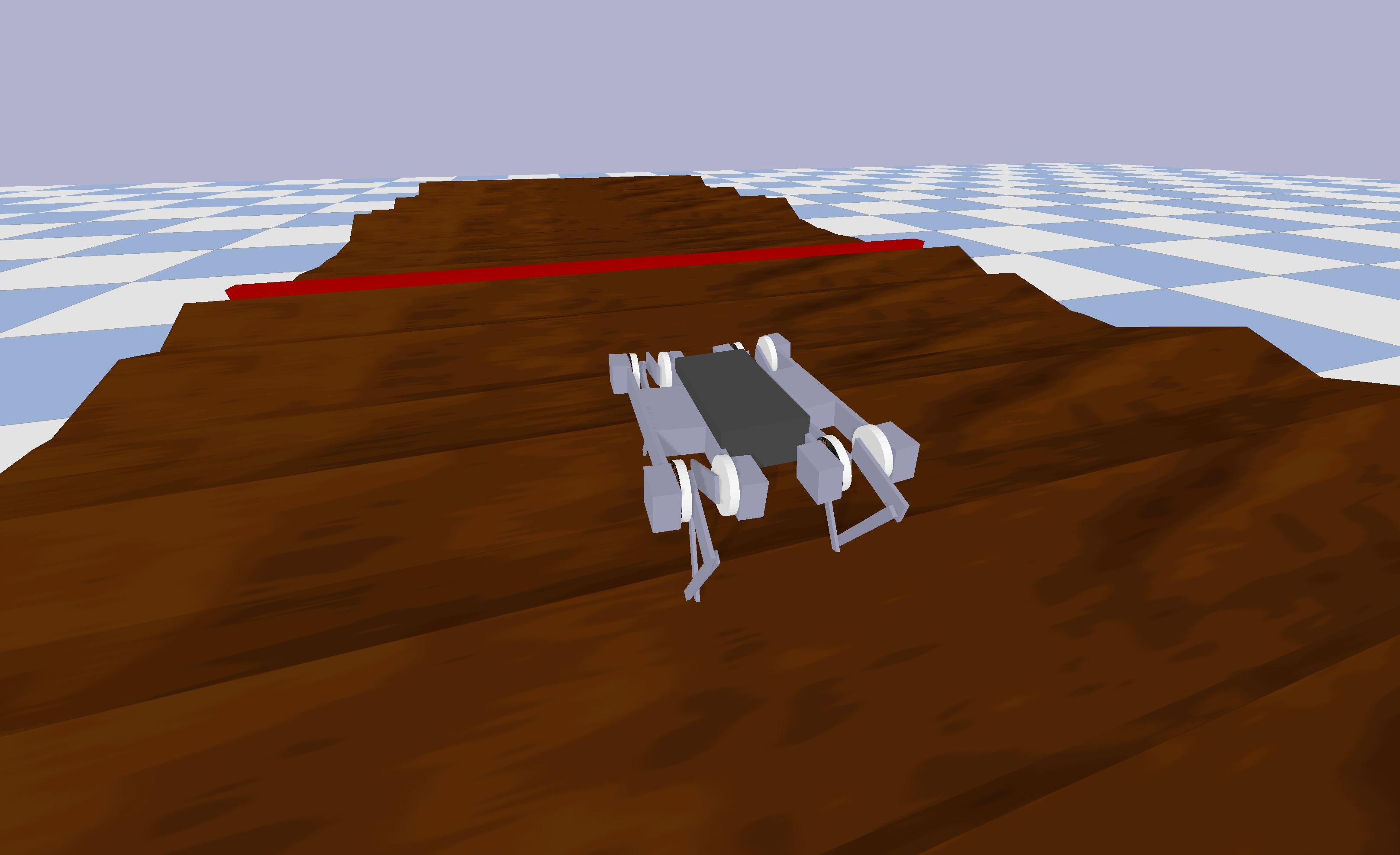}
\label{fig:minitaur}
}
\vskip -10pt
\caption{\small{Scenarios for vision-based planning. \textbf{(a)} A UAV that has to navigate across obstacle fields using onboard vision sensors and no prior knowledge of the environment map.  \textbf{(b)} A quadrupedal robot that has to traverse rough terrains with onboard vision sensors and proprioceptive feedback.}}
\vspace{-2mm}
\end{figure*}

Imagine an unmanned aerial vehicle (UAV) navigating through a dense environment using an RGB-D sensor (Figure \ref{fig:quadrotor}). Can we quantify the UAV's ability to successfully navigate across a previously unseen forest of obstacles without any collisions? The answer to this question can be crucial for various safety- or mission-critical applications. Various recent approaches for vision-based planning have sought to harness the power of deep learning for its ability to exploit statistical regularities and forego explicit geometric representations of the environment; a non-exhaustive list includes \citep{Bojarski16,Srinivas18,Ichter19}. However, such approaches are \emph{unable to provide explicit guarantees on the generalization performance} of the robot. In other words, they lack bounds on the performance of the learned planner when placed in a \emph{novel environment} (i.e., an environment that was not seen during training).

The goal of this paper is to address this challenge by developing an approach that learns to plan using vision sensors while providing \emph{explicit bounds on generalization performance}. In particular, the guarantees associated with our approach take the form of \emph{probably approximately correct} (PAC) \citep{McAllester99} bounds on the performance of the learned vision-based planner in novel environments. Concretely, given a set of training environments, we learn a planner with a provable bound on its expected performance in novel environments (e.g., a bound on the probability of collision). This bound holds with high probability (over sampled environments) under the assumption that training environments and novel environments are drawn from the same (but \emph{unknown}) underlying distribution (see Section \ref{sec:prob-form} for a formal description of our problem formulation). 

{\bf Statement of Contributions.} To our knowledge, the results in this paper constitute the first attempt to provide guarantees on generalization performance for vision-based planning using neural networks. To this end, we make three primary contributions. First, we develop a framework that leverages \emph{PAC-Bayes generalization theory} \citep{McAllester99} for learning to plan in a receding-horizon manner using a library of motion primitives. Switching among members of a motion primitive library facilitates the training by allowing us to leverage prior knowledge of the robot's low-level dynamics. The planners trained using this framework are accompanied by certificates of performance on novel environments in the form of PAC bounds. Second, we present algorithms based on \emph{evolutionary strategies} \citep{Wierstra2014} and \emph{convex optimization} (\emph{relative entropy programming} \citep{Chandrasekaran17}) for learning to plan with high-dimensional sensory feedback (e.g., RGB-D) by explicitly optimizing the PAC-Bayes bounds on the performance. Finally, we demonstrate the ability of our approach to provide strong generalization bounds for vision-based motion planners through two examples: navigation of a UAV across obstacle fields (see Fig.~\ref{fig:quadrotor}) and locomotion of a quadruped across rough terrains (see Fig.~\ref{fig:minitaur}). In both examples, we obtained PAC-Bayes bounds that guarantee successful traversal of $75-80\%$ of the environments on average.

\subsection{Related Work}
\label{subsec:lit}

\textbf{Planning with Motion Primitive Libraries.} Motion primitive libraries consist of pre-computed primitive trajectories that can be sequentially composed to generate a rich class of motions. The use of such libraries is prevalent in the literature on motion planning; a non-exhaustive list of examples includes navigation of UAVs \citep{Majumdar17}, balancing \citep{Liu09} and navigation \citep{Motahar16} of humanoids, grasping \citep{Berenson07}, and navigation of autonomous ground vehicles \citep{Sermanet08}. Several approaches furnish theoretical guarantees on the composition of primitive motions, such as: manuever automata \citep{Frazzoli05}, composition of funnel libraries by estimating regions-of-attraction \citep{Burridge99,Tedrake10,Majumdar17}, and leveraging the theory of switched systems \citep{Veer19}. However, unlike the present paper, none of the above provide theoretical guarantees for \emph{vision-based} planning by composing motion primitives.

\textbf{Planning with Vision.} The advent of Deep Neural Networks (DNNs) and Convolutional Neural Networks (CNNs) have recently boosted interest in learning vision-based motion planners. Certain recent methods can be classified in one of three categories: (1) self-supervised learning approaches \citep{Watter15,Ichter19} that uncover low-dimensional latent-space representations of the visual data before planning; (2) imitation learning approaches \citep{Bojarski16,Srinivas18,Qureshi19} that leverage an expert's motion plans; and (3) deep reinforcement learning (RL) \citep{Zhu17,Ebert18,Ha18,Lee19} that performs RL with visual data and uncovers the latent-space representations relevant to the planning task. Guarantees on planners presented in the above papers are limited to the learned low-dimensional latent space embeddings and do not necessarily translate to the actual execution of the plan. In this paper, we provide generalization guarantees for the \emph{execution} of our neural network planning policies in novel environments.

\textbf{Generalization Guarantees on Motion Plans.} 
Generalization theory was developed in the context of supervised learning to provide bounds on a learned model's performance on novel data \citep{Shalev14}. In the domain of robotics, these PAC generalization bounds were used in \citep{Karydis15} to learn a stochastic robot model from experimental data. PAC bounds were also adopted by the controls community to learn robust controllers \citep{Vidyasagar01,Campi19}; however, their use has not extended to vision-based DNN controllers/policies. In this paper, we use the PAC-Bayes framework, which has recently been successful in providing generalization bounds for DNNs for supervised learning \citep{Dziugaite17,Rivasplata19}. The PAC-Bayes Control framework was developed in \citep{Majumdar18,Majumdar19} to provide generalization bounds on learned control policies. Our paper differs from this work in three main ways. First, we \emph{plan} using motion primitive libraries instead of employing reactive control policies. This allows us to embed our knowledge of the robot's dynamics while simultaneously reducing the complexity of the policies. Second, we perform deep RL with PAC-Bayes guarantees using rich sensory feedback (e.g., $50\times 50$ depth map) and DNN policies. Finally, this paper contributes various algorithmic developments. We develop a training pipeline using Evolutionary Strategies (ES) \citep{Wierstra2014} to obtain a prior for the PAC-Bayes optimization. Furthermore, we develop an efficient Relative Entropy Program (REP)-based PAC-Bayes optimization with the recent quadratic-PAC-Bayes bound \citep[Theorem~1]{Rivasplata19} that was shown to be tighter than the bound used in \citep{Majumdar18,Majumdar19}. 

\section{Problem Formulation}
\label{sec:prob-form}

We consider robotic systems with discrete-time dynamics:
\begin{equation}\label{eq:rob-dyn}
x(t+1) = f(x(t),u(t); E) \enspace,
\end{equation}
where $t \geq 0$ is the time-step, $x(t) \in \X \subset \RR^{\mathrm{dim}(x)}$ is the robot's state, $u(t) \in \U \subset \RR^{\mathrm{dim}(u)}$ is the control input, and $E\in\mathcal{E}$ is the robot's ``environment". We use the term ``environment" broadly to refer to any exogenous effects that influence the evolution of the robot's state; e.g., the geometry of an obstacle environment that a UAV must navigate or the geometry of terrain that a legged robot must traverse. In this paper we will make the following assumption.
\begin{assumption}
\label{ass:D-sampling}
There is an underlying \emph{unknown} distribution $\mathcal{D}$ over the space $\mathcal{E}$ of all environments that the robot may be deployed in. At training time, we are provided with a dataset $S := \{E_1, E_2, \cdots, E_N\}$ of $N$ environments drawn i.i.d. from $\mathcal{D}$.
\end{assumption}
\vspace{-2.5mm}
It is important to emphasize that we \emph{do not} assume any explicit characterization of $\mathcal{D}$ or $\mathcal{E}$. We only assume indirect access to $\mathcal{D}$ in the form of a training dataset (e.g., a dataset of building geometries for the problem of UAV navigation).  

Let $g:\mathcal{X}\times\mathcal{E}\to\mathcal{O}$ be the robot's extereoceptive sensor (e.g., vision) that furnishes an observation $o\in\mathcal{O}$ from a state $x\in\mathcal{X}$ and an environment $E\in\mathcal{E}$. Further, let $h:\mathcal{X}\to\mathcal{Y}$ be the robot's proprioceptive sensor mapping that maps the robot's state $x\in\mathcal{X}$ to a sensor output $y\in\mathcal{Y}$. We aim to learn control policies that have a notion of \emph{planning} embedded in them. In particular, we will work with policies $\pi:\mathcal{O}\to\mathcal{L}$ that utilize rich sensory observations $\mathcal{O}$, e.g., vision or depth, to plan the execution of a motion primitive from a library $\mathcal{L}:=\{\Gamma_j:\mathbb{R}_{\geq 0}\times\mathcal{Y} \rightarrow \U~|~j\in\mathcal{J}\}$ in a \emph{receding-horizon} manner. Each member of $\mathcal{L}$ is a (potentially time-varying) proprioceptive controller and the index set $\mathcal{J}$ is compact. 

We assume the availability of a cost function that defines the robot's task. For the sake of simplicity, we will assume that the environment $E$ captures all sources of stochasticity (including random initial conditions); thus, the cost $C(\pi; E)$ associated with deploying policy $\pi$ on a \emph{particular} environment $E$ (over a given time horizon $T$) is deterministic. In order to apply PAC-Bayes theory, we assume that the cost $C(\pi; E)$ is bounded. Without further loss of generality, we assume $C(\pi; E) \in [0,1]$. As an example in the context of navigation, the cost function may assign a cost of 1 for colliding with an obstacle in a given environment (during a finite time horizon) and a cost of 0 otherwise. 

The goal of this work is to learn policies that minimize the expected cost across \emph{novel} environments drawn from $\mathcal{D}$:
\begin{equation}
\label{eq:opt_det}
\underset{\pi \in \Pi}{\textrm{min}} \  \ C_\mathcal{D}(\pi) :=  \underset{\pi \in \Pi}{\textrm{min}} \ \  \underset{E \sim \mathcal{D}}{\EE} \ [C(\pi; E)].
\end{equation}
As the distribution $\mathcal{D}$ over environments is \emph{unknown}, a direct computation of $C_{\mathcal{D}}(\pi)$ for the purpose of the minimization in \eqref{eq:opt_det} is infeasible. The PAC-Bayes framework \citep{Maurer04,McAllester99} provides us an avenue to alleviate this problem. 
However, in order to leverage it, we will work with a slightly more general problem formulation. In particular, we learn a \emph{distribution} $P$ over the space $\Pi$ of policies instead of finding a single policy. When the robot is faced with a given environment, it first randomly selects a policy using $P$ and then executes this policy. The corresponding optimization problem is:
\begin{flalign*}
&& C^\star := \underset{P \in \mathcal{P}}{\textrm{min}} \  \ C_\mathcal{D}(P) := \underset{P \in \mathcal{P}}{\textrm{min}} \  \  \underset{E \sim \mathcal{D}}{\EE} \ \underset{\pi \sim P}{\EE} [C(\pi; E)], &&& (\OPT)
\end{flalign*}
where $\mathcal{P}$ is the space of probability distributions over $\Pi$. We emphasize that the distribution $\mathcal{D}$ over environments is unknown to us. We are only provided a finite training dataset $S$ to learn from; solving $\OPT$ thus requires finding (distributions over) policies that \emph{generalize} to novel environments.

\section{PAC-Bayes Control}
\label{sec:pac-bayes-control}


We now extend the PAC-Bayes Control approach developed in \citep{Majumdar18, Majumdar19} to vision-based planning using motion primitives. Let $\Pi = \{\pi_w | w \in \RR^d\}$ denote a space of policies parameterized by weight vectors $w \in \RR^d$ that determine the mapping from observations in $\mathcal{O}$ to primitives in $\mathcal{L}$. Specifically, the parameters $w$ will correspond to weights of a neural network. Let $P_0$ represent a ``prior" distribution over control policies obtained by specifying a distribution over the parameter space $\RR^d$. The PAC-Bayes approach requires this prior to be chosen independently of the dataset $S$ of training environments. As described in Section \ref{sec:prob-form}, our goal is to learn a distribution $P$ over policies that minimizes the objective in $\OPT$. We will refer to $P$ as the ``posterior". We note that the prior and the posterior need not be Bayesian. We define the \emph{empirical cost} associated with a particular choice of posterior as the average (expected) cost across training environments in $S$:
\begin{equation}\label{eq:emp-cost}
C_S(P) := \frac{1}{N} \sum_{E \in S} \underset{w \sim P}{\mathbb{E}} [C(\pi_w; E)].
\end{equation}
The following theorem is the primary theoretical result we leverage in this work. 
\begin{theorem}\label{thm:mcallester-pac}
For any $\delta\in(0,1)$ and posterior $P$, with probability $1-\delta$ over sampled environments $S\sim \mathcal{D}^N$, the following inequalities hold:
\begin{align}
\text{(i)}~C_{\mathcal{D}}(P) & \leq C_{PAC}(P,P_0) := C_S(P) + \sqrt{R(P,P_0)} ,~~~~~~~~~~~~~~~~~~~~~~~~~~~~~~~~~~~~~~~~~~~~~~~~~~~~~~~~~~~~~ \label{eq:mcallester-pac-bound} \\
\text{(ii)}~C_{\mathcal{D}}(P) & \leq C_{QPAC}(P,P_0) := \big(\sqrt{C_S(P) + R(P,P_0)} + \sqrt{R(P,P_0)}\big)^2 , \label{eq:quad-pac-bound}
\end{align}
where
$R(P,P_0)$ is defined as:
\vspace{-2.5mm}
\begin{equation}\label{eq:R}
R(P,P_0):=\frac{\mathrm{KL}(P||P_0) + \log\big(\frac{2\sqrt{N}}{\delta}\big) }{2N} \enspace.
\end{equation} 
\end{theorem}
\vspace{-3mm}
\begin{proof}
\emph{(i)} was proved in \citep[Theorem~2]{Majumdar18}. The proof of \emph{(ii)} follows analogous to \citep[Theorem~2]{Majumdar18} with the only difference being that we use \citep[Theorem~1]{Rivasplata19} in the place of \citep[Corollary~1]{Majumdar18}.
\end{proof}
\vspace{-2.5mm}

This result provides an upper bound (that holds with probability $1 - \delta$) on our primary quantity of interest: the objective $C_\mathcal{D}(P)$ in $\OPT$. In other words, it allows us to bound the true expected cost of a posterior policy distribution $P$ across environments drawn from the (unknown) distribution $\mathcal{D}$. Theorem \ref{thm:mcallester-pac} suggests an approach for choosing a posterior $P$ over policies; specifically, one should choose a posterior that minimizes the bounds on $C_\mathcal{D}(P)$. The bounds are a composite of two quantities: the empirical cost $C_S(P)$ and a ``regularization" term $R(P)$ (both of which can be computed given the training dataset $S$ and a prior $P_0$). Intuitively, minimizing these bounds corresponds to minimizing a combination of the empirical cost and a regularizer that prevents overfitting to the specific training environments. 

For solving $\OPT$, we can either minimize \eqref{eq:mcallester-pac-bound} or \eqref{eq:quad-pac-bound}. Intuitively, we would like to use the tighter one of the two. The following proposition addresses this concern by analytically identifying regimes where \eqref{eq:mcallester-pac-bound} is tighter than \eqref{eq:quad-pac-bound} and vice-versa.

\begin{prop}\label{prop:pac-comp}
Let $C_{PAC}(P,P_0)$ and $C_{QPAC}(P,P_0)$ be the upper bounds of \eqref{eq:mcallester-pac-bound} and \eqref{eq:quad-pac-bound} respectively. Then, for any $P,P_0\in\mathcal{P}$ such that $\mathrm{KL}(P||P_0)<\infty$, the following hold:\footnote{For notational convenience we are dropping the dependence on $P$, $P_0$.}
\begin{enumerate}[{(i)}]
\vspace{-2.5mm}
\item $C_{QPAC} \leq 1/4 \iff C_{QPAC} \leq C_{PAC}$,
\vspace{-2.5mm}
\item $C_{QPAC} \geq 1/4 \iff C_{QPAC} \geq C_{PAC}$,
\vspace{-2.5mm}
\item $C_{QPAC} = 1/4 \iff C_{QPAC} = C_{PAC}$.
\vspace{-2.5mm}
\end{enumerate}
\end{prop}
The proof of this proposition is detailed in Appendix~\ref{app:proofs}.

Proposition~\ref{prop:pac-comp} shows that \eqref{eq:quad-pac-bound} is tighter than \eqref{eq:mcallester-pac-bound} if and only if the upper bound of \eqref{eq:quad-pac-bound} is smaller than $1/4$. On the other hand, we also have that \eqref{eq:mcallester-pac-bound} is tighter than \eqref{eq:quad-pac-bound} if and only if the upper bound of \eqref{eq:quad-pac-bound} is greater than $1/4$. Hence, in our PAC-Bayes training algorithm we use \eqref{eq:mcallester-pac-bound} when $C_{QPAC}>1/4$ and \eqref{eq:quad-pac-bound} when $C_{QPAC}<1/4$.

\section{Training}
\label{sec:train}

In this section we present our methodology for training vision-based planning policies that can provably perform well on novel environments using the PAC-Bayes Control framework. The PAC-Bayes framework permits the use of any prior distribution $P_0$ (independent of the training data) on the policy space. However, an uninformed choice of $P_0$ could result in vacuous bounds \citep{Dziugaite17}. Therefore, obtaining strong PAC-Bayes bounds with efficient sample complexity calls for a good prior $P_0$ on the policy space. For DNNs, the choice of a good prior is often unintuitive. To remedy this, we split a given training dataset into two parts: $\hat{S}$ and $S$. We use the Evolutionary Strategies (ES) framework to train a prior $P_0$ using the training data in $\hat{S}$; more details are provided in Section~\ref{subsec:train-ES}. Leveraging this prior, we perform PAC-Bayes optimization on the training data in $S$; further details on the PAC-Bayes optimization are presented in Section~\ref{subsec:pac-bayes-quad}.

\subsection{Training A PAC-Bayes Prior With ES}
\label{subsec:train-ES}

We train the prior distribution $P_0$ on the policy space $\Pi$ by minimizing the \emph{empirical cost} on environments belonging to the set $\hat{S}$ with cardinality $\hat{N}$. In particular, we choose $P_0$ to be a multivariate Gaussian distribution $\mathcal{N}(\mu, \Sigma)$ with a mean $\mu\in\mathbb{R}^d$ and a diagonal covariance $\Sigma\in\mathbb{R}^{d\times d}$. Let $\sigma\in\mathbb{R}^d$ be the element-wise square-root of the diagonal of $\Sigma$. Our training is performed using the class of RL algorithms known as Evolutionary Strategies (ES) \citep{Wierstra2014}. ES provides multiple benefits in our setting: (a) The presence of a physics engine in the training loop prohibits backpropagation of analytical gradients. Since the policies $\pi\in\Pi$ in our setting are DNNs with hundreds and thousands of parameters, a naive finite-difference estimation of the gradient is computationally prohibitive. ES permits gradient estimation with significantly lower number of rollouts (albeit resulting in noisy estimates).
(b) ES directly supplies us a distribution on the policy space, thereby meshing well with the PAC-Bayes Control framework. (c) The ES gradient estimation can be conveniently parallelized in order to leverage cloud computing resources.

Adapting \eqref{eq:emp-cost} for $\hat{S}$, we can express the gradient of the empirical cost with respect to (w.r.t.) $\psi:=(\mu,\sigma)\in\mathbb{R}^{2d}$ as:
\begin{equation}\label{eq:emp-grad-psi}
\nabla_\psi C_{\hat{S}}(P_0) = \frac{1}{\hat{N}} \sum_{E \in \hat{S}} \nabla_\psi\underset{w \sim P_0}{\mathbb{E}} [C(\pi_w; E)] \enspace.
\end{equation}
Following the ES framework from \citep{Salimans17,Wierstra2014}, the gradient of the empirical cost for any $E\in\hat{S}$ w.r.t. the mean $\mu$ and standard deviation $\sigma$ is:
\begin{align}
\nabla_\mu \underset{w \sim P_0}{\mathbb{E}} [C(\pi_w; E)] & = \underset{\epsilon\sim \mathcal{N}(0,I)}{\mathbb{E}} [C(\pi_{\mu+\sigma\odot\epsilon};E) \epsilon]\oslash \sigma \enspace, \label{eq:grad-mu-ES} \\
\nabla_\sigma \underset{w \sim P_0}{\mathbb{E}} [C(\pi_w; E)] & = \underset{\epsilon\sim \mathcal{N}(0,I)}{\mathbb{E}} [C(\pi_{\mu+\sigma\odot\epsilon};E) (\epsilon\odot\epsilon-\mathbf{1})] \oslash \sigma,\label{eq:grad-sigma-ES}
\end{align}
where $\oslash$ is the Hadamard division (element-wise division), $\odot$ is the Hadamard product (element-wise product), and $\mathbf{1}$ is a vector of $1$'s with dimension $d$.
One can derive \eqref{eq:grad-mu-ES} and \eqref{eq:grad-sigma-ES} using the diagonal covariance structure of $\Sigma$ and the reparameterization trick\footnote{$w\sim \mathcal{N}(\mu,\Sigma)\iff w = \mu + \sigma\odot\epsilon$ where $\epsilon\sim\mathcal{N}(0,I)$} in \citep[Equations~(3),(4)]{Wierstra2014}.
To reduce the variance in the gradient estimate we perform antithetic sampling, as suggested in \citep{Salimans17}, i.e., for every $\epsilon\sim\mathcal{N}(0,I)$ we also evaluate the policy corresponding to $-\epsilon$. If we sample $\hat{m}$ number of $\epsilon$, then the Monte-Carlo estimate of the gradient with antithetic sampling  is:
\begin{align}
\nabla_\mu \underset{w \sim P_0}{\mathbb{E}} [C(\pi_w; E)] & \approx \frac{1}{2\hat{m}}\sum_{i=1}^{2\hat{m}} [C(\pi_{\mu+\sigma\odot\epsilon_i};E) \epsilon]\oslash \sigma , \label{eq:grad-mu-ES-baseline-MC} \\
\nabla_\sigma \underset{w \sim P_0}{\mathbb{E}} [C(\pi_w; E)] & \approx \frac{1}{2\hat{m}}\sum_{i=1}^{2\hat{m}} [C(\pi_{\mu+\sigma\odot\epsilon_i};E) (\epsilon\odot\epsilon-\mathbf{1})] \oslash \sigma. \label{eq:grad-sigma-ES-baseline-MC}
\end{align}
These estimated gradients are then passed to a gradient-based optimizer for updating the distribution; we use the Adam optimizer \citep{Kingma14}. Appendix~\ref{app:ES} provides the psuedo-code for training a PAC-Bayes prior with ES and further implementation details.

\subsection{Training a PAC-Bayes Policy}
\label{subsec:pac-bayes-quad}

This section details our approach for minimizing the PAC-Bayes upper-bounds in Theorem~\ref{thm:mcallester-pac} to obtain provably generalizable posterior distributions on the policy-space. We begin by restricting our policy space to a finite set as follows:

\noindent \emph{Let $P_0$ be a prior on the policy space $\Pi$ obtained using ES. Draw $m$ i.i.d. policies $\{\pi_i\sim P_0~|~i=1,\cdots,m\}$ from $P_0$ and restrict the policy space to $\tilde{\Pi}:=\{\pi_i~|~ i=1,\cdots,m\}$. Finally, define a prior $p_0\in\mathbb{R}^m$ over the finite space $\tilde{\Pi}$ as a uniform distribution.}

The primary benefit of working over a finite policy space is that it allows us to formulate the problem of minimizing the PAC-Bayes bounds \eqref{eq:mcallester-pac-bound} and \eqref{eq:quad-pac-bound} using \emph{convex optimization}. As described in \citep[Section~5.1]{Majumdar19}, optimization of the PAC-Bayes bound \eqref{eq:mcallester-pac-bound} for $\tilde{\Pi}$ can be achieved using a \emph{relative entropy program (REP)}; REPs are efficiently-solvable convex programs in which a linear functional of the decision variables is minimized subject to constraints that are linear or of the form $\text{KL}(\cdot||\cdot)\leq c$ \citep[Section~1.1]{Chandrasekaran17}. The remainder of this section will formulate the optimization of the bound \eqref{eq:quad-pac-bound} as a parametric REP. The resulting algorithm (Algorithm~\ref{alg:pac-bayes}) finds a posterior distribution $p$ over $\tilde{\Pi}$ that minimizes the PAC-Bayes bound (arbitrarily close to the global infimum). 

Let $C\in\mathbb{R}^m$ be the policy-wise cost vector, each entry of which holds the average cost of running policy $\pi_i \in \tilde{\Pi}$ on the environments $\{E_i\}_{i=1}^N$. Then, the empirical cost $C_S(p)$ can be expressed linearly in $p$ as $Cp$. Hence, the minimization of the PAC-Bayes bound \eqref{eq:quad-pac-bound} can be written as: 
\vspace{-2mm}
\begin{align}\label{eq:REP-original}
\min_{p\in\mathbb{R}^m} ~ \big(\sqrt{Cp + R(p,p_0)} + \sqrt{R(p,p_0)}\big)^2 \quad \textrm{s.t.} ~ \sum_{i=1}^m p_i = 1, 0\leq p_i \leq 1.
\vspace{-3mm}
\end{align}
Introducing the scalars $\tau$, $\lambda$, and $\hat{C}$, and performing the epigraph trick \cite{Boyd04} -- detailed in Appendix~\ref{app:epigraph} -- this optimization can be equivalently expressed as:\footnote{If we have an infeasible $(\hat{C},\lambda)$, then we assume that $\mathcal{R}(\hat{C},\lambda)=\infty$.}
\begin{numcases}{\mathcal{R}(\hat{C},\lambda):}
\min_{p\in\mathbb{R}^m,\lambda\in\mathbb{R},\hat{C}\in\mathbb{R},\tau\in\mathbb{R}} \quad \tau \nonumber \\
~\textrm{s.t.} \quad \tau \geq \hat{C} + 2R(p,p_0) + 2\lambda, \label{eq:REP-ctr-1}\\
~\quad \quad \lambda^2 \geq \hat{C}R(p,p_0)+R(p,p_0)^2, \label{eq:REP-ctr-2}\\
~\quad \quad \hat{C} = Cp, \label{eq:REP-ctr-3}\\
~\quad \quad \sum_{i=1}^m p_i = 1, 0\leq p_i \leq 1. \label{eq:REP-ctr-4}
\end{numcases}
This is an REP for a fixed $\lambda$ and $\hat{C}$. Hence, we can perform a grid search on these scalars and solve $\mathcal{R}$ for each fixed tuple of parameters $(\hat{C},\lambda)$. We can control the density of the grid on $(\hat{C},\lambda)$ to get arbitrarily close to:
\vspace{-2.5mm}
\begin{equation}\label{eq:tau-inf}
\tau^* := \inf \{ \mathcal{R}(\hat{C},\lambda)~|~(\hat{C},\lambda)\in[0,1]\times[0,\infty) \}.
\end{equation}
The search space for $(\hat{C},\lambda)$ is impractically large in \eqref{eq:tau-inf}. The following Proposition remedies this by identifying compact intervals of $\lambda$ and $\hat{C}$ which contain $\tau^*$, thereby drastically shrinking their search space and promoting efficient solution of \eqref{eq:tau-inf} to arbitrary precision. 

\begin{prop}\label{prop:REP}
Consider the optimization problem \eqref{eq:tau-inf}. Let $\gamma>0$ and assume that $C_{QPAC}(p_0,p_0)\leq\gamma$. Let $C_{\min}$ and $C_{\max}$ be the minimum and maximum entries of $C$, respectively. Further, let $\lambda_{\min}:=\sqrt{C_{\min}R(p_0,p_0)+R(p_0,p_0)^2}$ and $\lambda_{\max}:=(\gamma-C_{\min})/2 - R(p_0,p_0)$. Then, $\tau^*$, defined in \eqref{eq:tau-inf}, satisfies:
\vspace{-1mm}
\begin{align*}
\tau^* = \inf \{ \mathcal{R}(\hat{C},\lambda)\,|\,(\hat{C},\lambda)\in[C_{\min},C_{\max}]\times[\lambda_{\min},\lambda_{\max}] \}.
\end{align*}
\end{prop}
\vspace{-2.5mm}
The proof of Proposition~\ref{prop:REP} is provided in Appendix~\ref{app:proofs}.

Our implementation of the PAC-Bayes optimization is detailed in Algorithm~\ref{alg:pac-bayes}. In practice, we sweep $\hat{C}$ across $[C_{\min},C_{\max}]$ and for each fixed $\hat{C}$ we perform a bisectional search on $\lambda\in[\lambda_{\min},\lambda_{\max}]$ (line 13-14 in Algorithm~\ref{alg:pac-bayes}). We make our bounds for $\lambda$ in Proposition~\ref{prop:REP} tighter by replacing $C_{\min}$ with the chosen $\hat{C}$ in the expressions of $\lambda_{\min}$ and $\lambda_{\max}$ (line 10-11 in Algorithm~\ref{alg:pac-bayes}). Furthermore, we choose $\gamma=C_{QPAC}(p_0,p_0)$ (line 6 in Algorithm~\ref{alg:pac-bayes}). The REP that arises by fixing $(\hat{C},\lambda)$ is solved using CVXPY \citep{CVXPY} with the MOSEK solver \citep{MOSEK}. 

\begin{algorithm}[t]
\caption{PAC-Bayes Optimization}
\label{alg:pac-bayes}
\small
\begin{algorithmic}[1]
	\Procedure{PAC-Bayes-Opt}{$\mu_0, \sigma_0, \{E_i\}_{i=1}^N$}
		\State Initialize: $\tilde{\Pi}\gets$ $m$ i.i.d. policies from $\mathcal{N}(\mu_0,\Sigma_0)$, ~ $p_0\in\mathbb{R}^m$ as the uniform distribution
		\State $p^* \gets p_0$, $\tau^*\gets C_{QPAC}(p_0,p_0)$ 
		\State Compute: cost vector $C\in\mathbb{R}^m$; each entry $C[j]$ is the average cost of policy $j$ on $\{E_i\}_{i=1}^N$
		\State $\gamma \gets C_{QPAC}(p_0,p_0)$; ~$C_{\min}\gets \min_j C[j]$; ~$C_{\max}\gets \max_j C[j]$
		\State $\hat{C}\gets$ $K$ equally spaced points in $[C_{\min}, C_{\max}]$.
		\For{$i=1,\cdots,K$}
			\State $\lambda_{\min}\gets \sqrt{\hat{C}[i]R(p_0,p_0)+R(p_0,p_0)^2}$
			\State $\lambda_{\max}\gets (\gamma-\hat{C}[i])/2 - R(p_0,p_0)$
			\If{$\lambda_{\min}<\lambda_{\max}$} \Comment{Else $\tau\geq\gamma$}
				\State \Comment{Bisectional search on $\lambda$:\quad\quad\quad\quad\quad\quad\quad\quad\quad\quad\quad\quad\quad\quad\quad\quad\quad\quad\quad\quad\quad\quad\quad\quad\quad\quad\quad\quad\quad}
				\State $(p,\tau)\gets$ $\min_{\lambda\in[\lambda_{\min},\lambda_{\max}]} \mathcal{R}(\hat{C}[i],\lambda)$
				\If{$\tau<\tau^*$} 
					\State $(p^*,\tau^*)\gets (p,\tau)$
				\EndIf
			\EndIf
		\EndFor\\
	\Return $(p^*,\tau^*)$
	\EndProcedure
	\end{algorithmic}
\normalsize
\end{algorithm}

\vspace{-0.4mm}
\section{Examples}
\label{sec:examples}

In this section we use the algorithms developed in Section~\ref{sec:train} on two examples: (1) vision-based navigation of a UAV in novel obstacle fields, and (2) locomotion of a quadrupedal robot across novel rough terrains using proprioceptive and exteroceptive sensing. Through these examples, we demonstrate the ability of our approach to train vision-based DNN policies with strong generalization guarantees in a deep RL setting. The simulation and training are performed with PyBullet \citep{Coumans18} and PyTorch \citep{PyTorch} respectively. Our code is available at: {\small \url{https://github.com/irom-lab/PAC-Vision-Planning}} and a video of the results is available at: {\small \url{https://youtu.be/03qq4sLU34o}}.

\subsection{Vision-Based UAV Navigation}
\label{subsec:quadrotor}
In this example we train a quadrotor to navigate across an obstacle field without collision using depth maps from an \emph{onboard} vision sensor; see Fig.~\ref{fig:quadrotor} for an illustration. 

\begin{table}[b]
  \centering
  \begin{adjustbox}{width=0.85\columnwidth,center}
  \small
  \begin{tabular}{|cccc|cccc|}
    \hline
     \multicolumn{4}{|c|}{\bf \rule{0pt}{2ex} UAV Navigation} &  \multicolumn{4}{c|}{\bf Quadrupedal Locomotion}\\
     \hline
    \multicolumn{1}{|c|}{\rule{0pt}{2ex}\# Envs}  & \multicolumn{2}{c}{PAC-Bayes Cost} & {True Cost} & \multicolumn{1}{c|}{\rule{0pt}{2ex}\# Envs}  & \multicolumn{2}{c}{PAC-Bayes Cost} & {True Cost} \\
    \multicolumn{1}{|c|}{\rule{0pt}{2ex}N} & $C_{PAC}$ & $C_{QPAC}$ & (Estimate) & \multicolumn{1}{c|}{\rule{0pt}{2ex}N} & $C_{PAC}$ & $C_{QPAC}$ & (Estimate) \\
	\hline    
    \multicolumn{1}{|c|}{\rule{0pt}{2ex}1000} & 26.02\% & -- & 18.34\% & \multicolumn{1}{c|}{\rule{0pt}{2ex}1000} & -- & 23.87\% & 18.31\%\\
    \multicolumn{1}{|c|}{2000} & -- & 22.99\% & 18.42\% & \multicolumn{1}{c|}{\rule{0pt}{2ex}2000} & -- & 21.93\% & 18.29\%\\
    \multicolumn{1}{|c|}{4000} & -- & 21.55\% & 18.43\% & \multicolumn{1}{c|}{\rule{0pt}{2ex}4000} & -- & 20.73\% & 17.81\%\\
    \hline
  \end{tabular}
  \normalsize
  \end{adjustbox}
  \caption{\small PAC-Bayes Cost for UAV Navigation and Quadrupedal Locomotion \label{tab:results}}
  \vspace{-4mm}
\end{table}

\textbf{Environment.} For the sake of visualization, we made the ``roof" of the obstacle course transparent in Fig.~\ref{fig:quadrotor} and the videos. The true obstacle course is a (red) tunnel cluttered by cylindrical obstacles; see Fig.~\ref{fig:quadrotor-rgb} in Appendix~\ref{app:examples}. The (unknown) distribution $\mathcal{D}$ over environments is chosen by drawing obstacle radii and locations from a uniform distribution on $[5\,\textrm{cm},30\,\textrm{cm}]$ and $[-5\,\textrm{m},5\,\textrm{m}]\times [0\,\textrm{m},14\,\textrm{m}]$, respectively. The orientation is generated by drawing a quaternion using a normal distribution.

\textbf{Motion Primitives.} We work with a library $\mathcal{L}$ of $25$ motion primitives for the quadrotor. 
\begin{wrapfigure}{r}{0.3\textwidth}
    \begin{center}
        \vspace{-5mm}
        \includegraphics[width=0.3\textwidth]{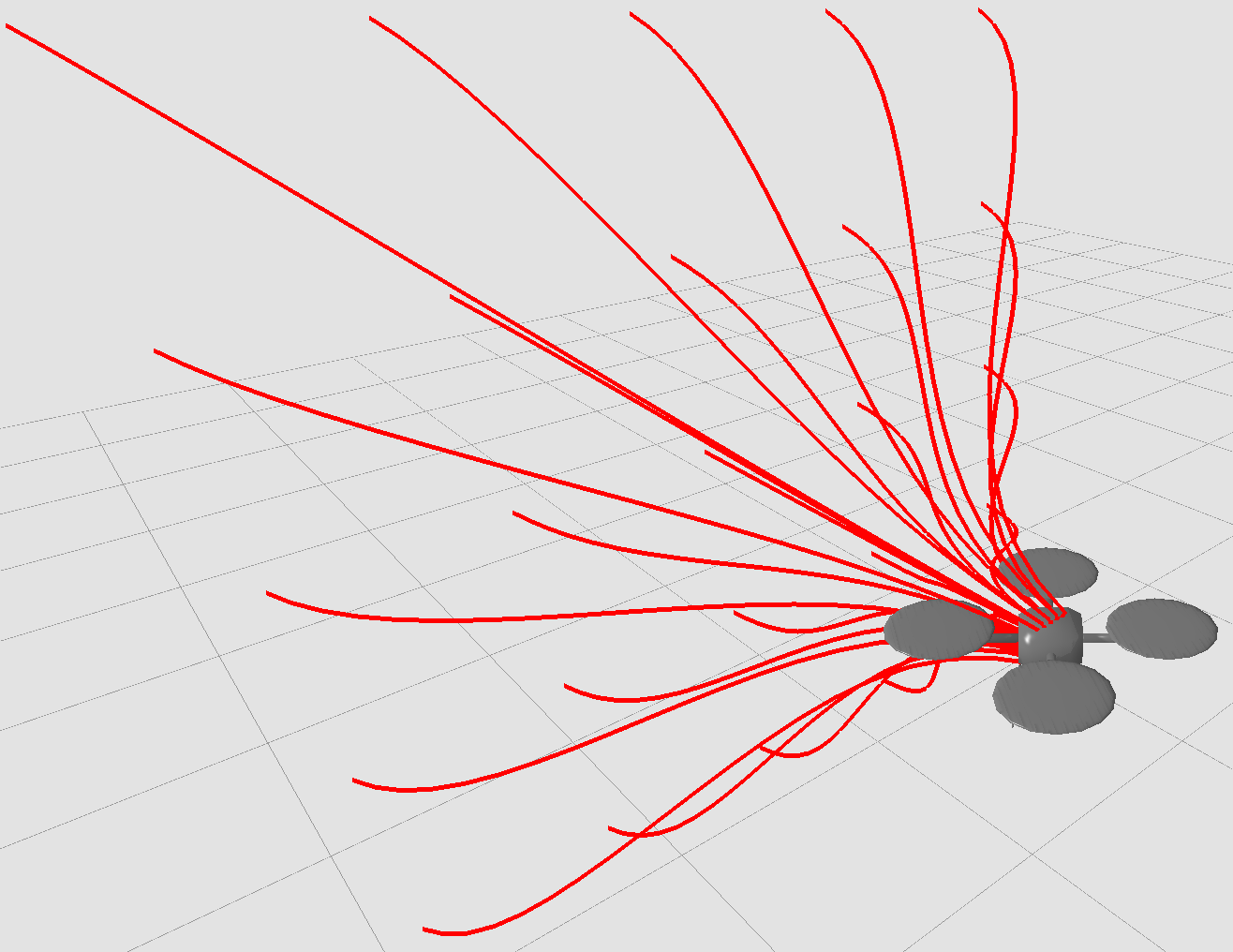}
    \end{center}
    \vspace{-3mm}
    \caption{\small Motion primitive trajectories for the quadrotor. \label{fig:quadrotor-prims}}
    \vspace{-5mm}
\end{wrapfigure}
The motion primitives are generated by connecting the initial position $(x_0,y_0,z_0)$ and the final desired position $(x_0+\Delta x,y_0+\Delta y,z_0+\Delta z)$ of the quadrotor by a smooth sigmoidal trajectory; Fig.~\ref{fig:quadrotor-prims} illustrates the sigmoidal trajectory for each primitive in our library. The robot moves along these trajectories at a constant speed and yaw and the roll-pitch are recovered by exploiting the differential flatness of the quadrotor \citep[Section~III]{Mellinger2011}.

\textbf{Planning Policy.} Our control policy $\pi:\mathcal{O}\to\mathcal{L}$ maps a depth map to a score vector and then selects the motion primitive with the highest score. Further details of the policy can be found in Appendix~\ref{app:UAV}.

\textbf{Training Summary.} We choose the cost as $1-t/T$ where $t$ is the time at which the robot collides with an obstacle and $T$ is the total time horizon; in our example $T=12$ seconds. For the primitives in Fig.~\ref{fig:quadrotor-prims}, the quadrotor moves monotonically forward. Hence, time is analogous to the forward displacement. The prior is trained using the method described in Section~\ref{subsec:train-ES} on an AWS EC2 {\tt g3.16xlarge} instance with a run-time of $\sim$22 hours. PAC-Bayes optimization is performed with Algorithm~\ref{alg:pac-bayes} on a desktop with a 3.30 GHz i9-7900X CPU with 10 cores, 32 GB RAM, and a 12 GB NVIDIA Titan XP GPU. The bulk of the time in executing Algorithm~\ref{alg:pac-bayes} is spent on computing the costs in line 5 (i.e., $\sim$400 sec, $\sim$800 sec, and $\sim$1600 sec for the results in Table~\ref{tab:results} from top to bottom), whereas solving \eqref{eq:tau-inf} takes $\sim 1$ sec; Table~\ref{tab:hyper} in Appendix~\ref{app:examples} supplies the exact hyperparameters.

\textbf{Results.} The PAC-Bayes results are detailed in Table~\ref{tab:results}. Here we choose $\delta=0.01$ (implying that the PAC-bounds hold with probability $0.99$) and vary $N$. We also perform an exhaustive simulation of the trained posterior on $5000$ novel environments to empirically estimate the true cost. It can be observed that our PAC-Bayes bounds show close proximity to the empirical estimate of the true cost. To facilitate the physical interpretation of the results in Table~\ref{tab:results}, consider the last row with $N=4000$: according to our PAC-Bayes guarantee, with probability $0.99$ the quadrotor will (on average) get through $78.45\%~(100\%-21.55\%)$ of previously unseen obstacle courses.

\subsection{Quadrupedal Locomotion on Uneven Terrain}
\label{subsec:minitaur}

\textbf{Environment.} In this example, we train the quadrupedal robot Minitaur \citep{Kenneally16} to traverse an uneven terrain characterized by slopes uniformly sampled between $0^{\rm o}$ to $35^{\rm o}$; see Fig.~\ref{fig:minitaur} for an illustration of a representative environment. We use the {\tt minitaur\_gym\_env} in PyBullet to simulate the full nonlinear/hybrid dynamics of the robot. The objective here is to train a posterior distribution on policies that enables the robot to cross a finish-line (depicted in red in Fig.~\ref{fig:minitaur}) situated $\tilde{x}$ m in the initial heading direction $X$.

\textbf{Motion Primitives.} We use the sine controller that is available in the {\tt minitaur\_gym\_env} as well as Minitaur's SDK developed by Ghost Robotics. The sine controller generates desired motor angles based on a sinusoidal function that depends on the stepping amplitudes $a_1$, $a_2$, steering amplitude $a_{\rm st}$, and angular velocity $\omega$ as follows: $u_1 = (a_1 + a_{\rm st}) \sin(\omega t); ~ u_2 = (a_1 - a_{\rm st}) \sin(\omega t + \pi); ~ u_3 = a_2 \sin(\omega t); ~ u_4 = a_2 \sin(\omega t + \pi).$
These four desired angles are communicated to the eight motors of the Minitaur (two per leg) in the following order: $[u_1, u_2, u_2, u_1, u_3, u_4, u_4, u_3]$. Hence, our primitives are characterized by the scalars $a_1,a_2,a_{\rm st}\in[0,1]$, and $\omega\in[20,40]$ rad/s, rendering the library of motion primitives $\mathcal{L}$ uncountable but compact. Each primitive is executed for a time-horizon of 0.5 seconds as it roughly corresponds to one step for the robot.

\textbf{Planning Policy.} The policy $\pi:\mathcal{O}\times\mathcal{Y}\to\mathcal{L}$ selects a primitive based on the depth map and the proprioceptive feedback (8 motor angles, 8 motor angular velocities, and the six-dimensional position and orientation of the robot's torso). Further details of the policy can be found in Appendix~\ref{app:minitaur}.

\textbf{Training Summary.} The cost we use is $1-\Delta x/\tilde{x}$ where $\Delta x$ is the robot's displacement at the end of the rollout along the initial heading direction $X$ and $\tilde{x}$ is the distance to the finish-line from the robot's initial position along $X$; $\tilde{x}=6$m for the results in this paper. A rollout is considered complete when the robot crosses the finish-line or if the rollout time exceeds $10$ seconds. All the training in this example is performed on a desktop with a 3.30 GHz i9-7900X CPU with 10 cores, 32 GB RAM, and a 12 GB NVIDIA Titan XP GPU. As before, we train the prior using the method described in Section~\ref{subsec:train-ES}, the execution of which takes $\sim$2 hours. PAC-Bayes optimization is performed using Algorithm~\ref{alg:pac-bayes}. The execution of line 5 in Algorithm~\ref{alg:pac-bayes} takes 
$\sim$130 min, $\sim$260 min, and $\sim$520 min for the results in Table~\ref{tab:results} from top to bottom, respectively, whereas solving \eqref{eq:tau-inf} takes $\sim$1 sec. Hyperparameters can be found in Table~\ref{tab:hyper} in Appendix~\ref{app:examples}.

\textbf{Results.} The PAC-Bayes results are detailed in Table~\ref{tab:results} where $\delta=0.01$ and the number of environments $N$ is varied. The PAC-Bayes cost can be interpreted in a manner similar to the quadrotor; e.g., for PAC-Bayes optimization with $N=4000$, with probability $0.99$, the quadruped will (on average) traverse $79.27\%$ ($100\%-20.73\%$) of the previously unseen environments.

\section{Conclusions}
\label{sec:conclusions}

We presented a deep reinforcement learning approach for synthesizing vision-based planners with \emph{certificates of performance} on novel environments. We achieved this by directly optimizing a PAC-Bayes generalization bound on the average cost of the policies over all environments. To obtain strong generalization bounds, we devised a two-step training pipeline. First, we use ES to train a good prior distribution on the space of policies. Then, we use this prior in a PAC-Bayes optimization to find a posterior that minimizes the PAC-Bayes bound. The PAC-Bayes optimization is formulated as a parametric REP that can be solved efficiently. Our examples demonstrate the ability of our approach to train DNN policies with strong generalization guarantees. 

\textbf{Future Work.} There are a number of exciting future directions that arise from this work. We believe that our approach can be extended to provide \emph{certificates of generalization} for \emph{long-horizon} vision-based motion plans by combining our PAC-Bayes framework with generative networks that can predict the future visual observations conditioned on the primitive to be executed. 
Another direction we are excited to pursue is training vision-based policies that are robust to test environments which are not drawn from the same distribution as the training environments; e.g., wind gusts which are not a part of the training data. It is worth pointing out that our framework, currently, requires the training and the test data to be sampled from the same distribution to provide generalization guarantees. We hope to address this challenge by bridging the approach in this paper with existing model-based robust planning approaches \citep{Majumdar17,Veer19} and robust PAC-Bayes bounds; e.g., Corollary~4 in \cite{Majumdar19}.
%
%
Finally, we are also working towards a hardware implementation of our approach on a UAV and the Minitaur; a description of our proposed experimental setup is provided in Appendix~\ref{app:expts}.
%

\acknowledgments{The authors were supported by the Office of Naval Research [Award Number: N00014-18-1-2873], NSF [IIS-1755038], the Google Faculty Research Award, and the Amazon Research Award.}

\bibliography{irom}

\clearpage

\appendix
\section*{\Large Appendix}

\section{Proofs}
\label{app:proofs}
\begin{proof}[{\bf Proof of Proposition~\ref{prop:pac-comp}}]
We will begin with (i)\footnote{For notational convenience we drop the dependence of $C_S$, $R$ on $P$.}
\small
\begin{align}
\hspace{-2.4mm} C_{QPAC} \leq \frac{1}{4} \hspace{-1mm} & \iff \big(\sqrt{C_S + R} + \sqrt{R}\big)^2 \leq \frac{1}{4} \\
& \iff 2\sqrt{C_S + R} + 2\sqrt{R} \leq 1 \label{eq:pac-comp-1}\\
& \iff 2\sqrt{R(C_S + R)} + 2R + C_S \leq \sqrt{R} + C_S \label{eq:pac-comp-2}\\
& \iff \big(\sqrt{C_S + R} + \sqrt{R}\big)^2 \leq \sqrt{R} + C_S \label{eq:pac-comp-3}\\
& \iff C_{QPAC} \leq C_{PAC} \enspace,
\end{align}
\normalsize
where the step from \eqref{eq:pac-comp-1} to \eqref{eq:pac-comp-2} follows by multiplying both sides with $\sqrt{R}$, noting that $\sqrt{R}\geq 0$, and adding $C_S$; and the step from \eqref{eq:pac-comp-2} to \eqref{eq:pac-comp-3} follows by noting that $\big(\sqrt{C_S + R} + \sqrt{R}\big)^2 = 2\sqrt{R(C_S + R)} + 2R + C_S$.

The statement (ii) holds because it is the contrapositive of (i). Finally, the proof of (iii) follows by noting that for $C_{QPAC}=1/4$ both (i) and (ii) hold simultaneously, i.e., $C_{QPAC} \leq C_{PAC}$ as well as $C_{QPAC} \geq C_{PAC}$.
\end{proof}

\begin{proof}[{\bf Proof of Proposition~\ref{prop:REP}}]
The bound $C_{\min}\leq \hat{C}\leq C_{\max}$ follows by noting that for any $\hat{C}$ outside that interval the constraints \eqref{eq:REP-ctr-3} and \eqref{eq:REP-ctr-4} will be violated.

Let $\tilde{\mathcal{P}}$ be the space of probability distributions on $\tilde{\Pi}$. To get the bound on $\lambda$ first observe that for all $p\in\tilde{\mathcal{P}}$:
\begin{equation}\label{eq:Rp-p0}
R(p,p_0)\geq R(p_0,p_0) \enspace.
\end{equation}
Now, using the above and $\hat{C}\geq C_{\min}$ in \eqref{eq:REP-ctr-2}, we get:
\begin{equation}\nonumber
\lambda \geq \sqrt{C_{\min}R(p_0,p_0)+R(p_0,p_0)^2}=:\lambda_{\min} .
\end{equation}
Furthermore, it can be verified that:

\vspace{-3mm}
\small
\begin{equation*}
C_{QPAC}(p_0,p_0) = \mathcal{R}(Cp_0,\sqrt{Cp_0R(p_0,p_0)+R(p_0,p_0)^2})
\end{equation*}
\normalsize
ensuring that $C_{QPAC}(p_0,p_0)\geq \tau^*$. This fact, coupled with the hypothesis $C_{QPAC}(p_0,p_0)\leq \gamma$ in the statement of the proposition implies that $\tau^*\leq C_{QPAC}(p_0,p_0)\leq \gamma$, thereby allowing us to eliminate any $(C,\lambda)$ tuple for which $\tau>\gamma$. Hence, using $\tau\leq \gamma$, \eqref{eq:Rp-p0}, and $\hat{C}\geq C_{\min}$ in \eqref{eq:REP-ctr-1} we get that $\lambda\leq (\gamma-C_{\min})/2 - R(p_0,p_0)=:\lambda_{\max}$.
\end{proof}

\section{Implementation Details for Evolutionary Strategies}
\label{app:ES}
We exploit the high parallelizability of ES by splitting the $\hat{N}$ environments $\{E_i\}_{i=1}^{\hat{N}}$ into $k$ mini-batches $\{\hat{E}_i\}_{i=1}^k$, where $k$ is the number of CPU workers available to us. Each worker $i$ computes the gradient of the cost associated with each environment in $\hat{E}_i$ using Algorithm~\ref{alg:ES-env} and returns their sum to the main process, where the gradients are averaged over all $N$ environments; see Algorithm~3. These estimated gradients are then passed to a gradient-based optimizer for updating the distribution; we use the Adam optimizer \citep{Kingma14}. We circumvent the non-negativeness constraint of the standard deviation by optimizing\footnote{The notation $\log$ is overloaded to mean element-wise $\log$.} $\log(\sigma\odot\sigma)$ instead of $\sigma$. Accordingly, the gradients w.r.t. $\sigma$ must be converted to gradients w.r.t. $\log(\sigma\odot\sigma)$ before supplying them to the optimizer. We empirically observed that using ES, the training sometimes gets stuck at a local minimum. To remedy this, we replaced the cost function with the utility function \citep[Section~3.1]{Wierstra2014} as was done in \citep{Salimans17}. We shall refer to the use of the utility function as ES-utility. The difference between our implementation and \citep{Salimans17} is that we use ES-utility only when we get stuck in a local minimum; once we escape it, we revert back to ES. However, \citep{Salimans17} uses ES-utility for the entire duration of the training. The switching strategy allows us to benefit from the faster convergence of ES and local minimum avoidance of ES-utility, thereby saving a significant amount of computation.

A complete implementation of the ES algorithm used in the paper for training the prior is supplied in Algorithm~\ref{alg:ES-whole}.

\begin{algorithm}[h]
\caption{Train Prior using ES \label{alg:ES-whole}}
\small
\begin{algorithmic}[1]
	\Procedure{Train-Prior}{$\mu, \sigma, \{E_i\}_{i=1}^{\hat{N}}$}
		\State Initialize: $\mu$, $\sigma$, \textsc{Optimizer}
	\Repeat
		\State Generate environment mini-batches: $\{\hat{E}_i\}_{i=1}^k$
		\State Initialize: $\nabla_\mu C\gets\mathbf{0}$, $\nabla_\sigma C\gets\mathbf{0}$, $k$ workers
		\For{worker $i=1,\cdots,k$}
			\State Initialize: $\nabla_\mu C_i\gets 0$, $\nabla_\sigma C_i\gets 0$
			\For{$j=1,\cdots,\text{len}(\hat{E}_i)$}
				\State $(\nabla_\mu C_E, \nabla_\sigma C_E)\gets$ \textsc{ES-Grad}($\mu, \sigma, E_i$)
				\State $\nabla_\mu C_i\gets\nabla_\mu C_i + \nabla_\mu C_E$ \
				\State $\nabla_\sigma C_i\gets\nabla_\sigma C_i + \nabla_\sigma C_E$ \
				\EndFor
			\State Communicate $\nabla_\mu C_i$, $\nabla_\sigma C_i$ to main process
			\EndFor
		\State $\nabla_\mu C\gets \frac{1}{\hat{N}}\sum_{i=1}^k \nabla_\mu C_i$\
		\State $\nabla_\sigma C\gets \frac{1}{\hat{N}}\sum_{i=1}^k \nabla_\sigma C_i$\
		\State $(\mu, \sigma) \gets$ \textsc{Optimizer}$(\nabla_\mu C, \nabla_\sigma C)$\
	\Until Termination conditions satisfied\\
	\Return $\mu$, $\sigma$
	\EndProcedure
	\end{algorithmic}
\normalsize
\end{algorithm}

\begin{algorithm}[h]
\caption{ES Gradient on one Environment \label{alg:ES-env}}
\small
\begin{algorithmic}[1]
	\Procedure{ES-Grad}{$\mu, \sigma, E$}
		\State Initialize: $costs\gets\emptyset$
		\State Sample $\epsilon_1,\cdots, \epsilon_{\hat{m}}\sim\mathcal{N}(0,I)$.\
		\For{$i=1,\cdots,\hat{m}$}
			\State Compute $C(\mu+\sigma\epsilon_i;E)$ and $C(\mu-\sigma\epsilon_i;E)$\
			\State Append $C(\mu+\sigma\epsilon_i;E)$ and $C(\mu-\sigma\epsilon_i;E)$ to $costs$.\
		\EndFor
		\State Compute $\nabla_\mu C_E$, $\nabla_\sigma C_E$ using \eqref{eq:grad-mu-ES-baseline-MC} and \eqref{eq:grad-sigma-ES-baseline-MC}, respectively, with $costs$ and $\epsilon_1, \cdots, \epsilon_{\hat{m}}$\\
	\Return $(\nabla_\mu C_E$, $\nabla_\sigma C_E)$
	\EndProcedure
	\end{algorithmic}
\normalsize
\end{algorithm}

\section{Equivalence of \eqref{eq:REP-original} and $\mathcal{R}$}
\label{app:epigraph}

Let $\tau\in\mathbb{R}$ be an upper bound on the objective function of \eqref{eq:REP-original}. Then, \eqref{eq:REP-original} can be expressed as:
\begin{numcases}{}
\min_{p\in\mathbb{R}^m,\tau\in\mathbb{R}} \quad \tau \nonumber \\
~\textrm{s.t.} \quad \tau \geq \big(\sqrt{Cp + R(p,p_0)} + \sqrt{R(p,p_0)}\big)^2 \\
~\phantom{s.t. \quad \tau} = Cp + 2R(p,p_0) + 2\sqrt{CpR(p,p_0)+R(p,p_0)^2}, \label{eq:REP-form-1}\\
~\quad \quad \sum_{i=1}^m p_i = 1, 0\leq p_i \leq 1. 
\end{numcases}

Now we introduce the scalar $\lambda\in\mathbb{R}$ that upper bounds the square-root term of \eqref{eq:REP-form-1} giving us:
\begin{numcases}{}
\min_{p\in\mathbb{R}^m,\tau\in\mathbb{R}} \quad \tau \nonumber \\
~\textrm{s.t.} \quad \tau \geq Cp + 2R(p,p_0) + 2\lambda, \\
~\quad \quad \lambda^2 \geq CpR(p,p_0)+R(p,p_0)^2 \\
~\quad \quad \sum_{i=1}^m p_i = 1, 0\leq p_i \leq 1. 
\end{numcases}

Finally, we introduce $\hat{C} = Cp$ in this optimization to obtain $\mathcal{R}$.

\section{Further Implementation Details for Examples}
\label{app:examples}

The hyperparameters used for the examples in this paper are detailed in Table~\ref{tab:hyper}.

\begin{table}[h]
  \caption{Hyperparameters for Training \label{tab:hyper}}
  \vspace{-2mm}
  \centering
  \begin{tabular}{|c|ccccc|c|}
  	\hline
     & \multicolumn{5}{c|}{Training for Prior} & PAC-Bayes Opt.\\
    Example & Initial & \# Envs. & \# $\epsilon$ & \multicolumn{2}{c|}{learning rate} & \# Policy \\
     & Dist. & $\hat{N}$	 & $\hat{m}$ & $\mu$ & $\log(\sigma\odot\sigma)$ & $m$ \\
    \hline
    UAV\rule{0pt}{2.5ex} & $\mathcal{N}(0,4I)$ & 480 & 50 & 1 & 0.01 & 50 \\
    Quadruped & $\mathcal{N}(0,4I)$ & 10 & 5 & 1 & 0.01 & 50 \\
    \hline
  \end{tabular}
  \vspace{-2mm}
\end{table}


\subsection{Vision-based UAV Navigation}
\label{app:UAV}

\begin{wrapfigure}{r}{0.4\textwidth}
    \begin{center}
        \vspace{-5mm}
        \includegraphics[width=0.4\textwidth]{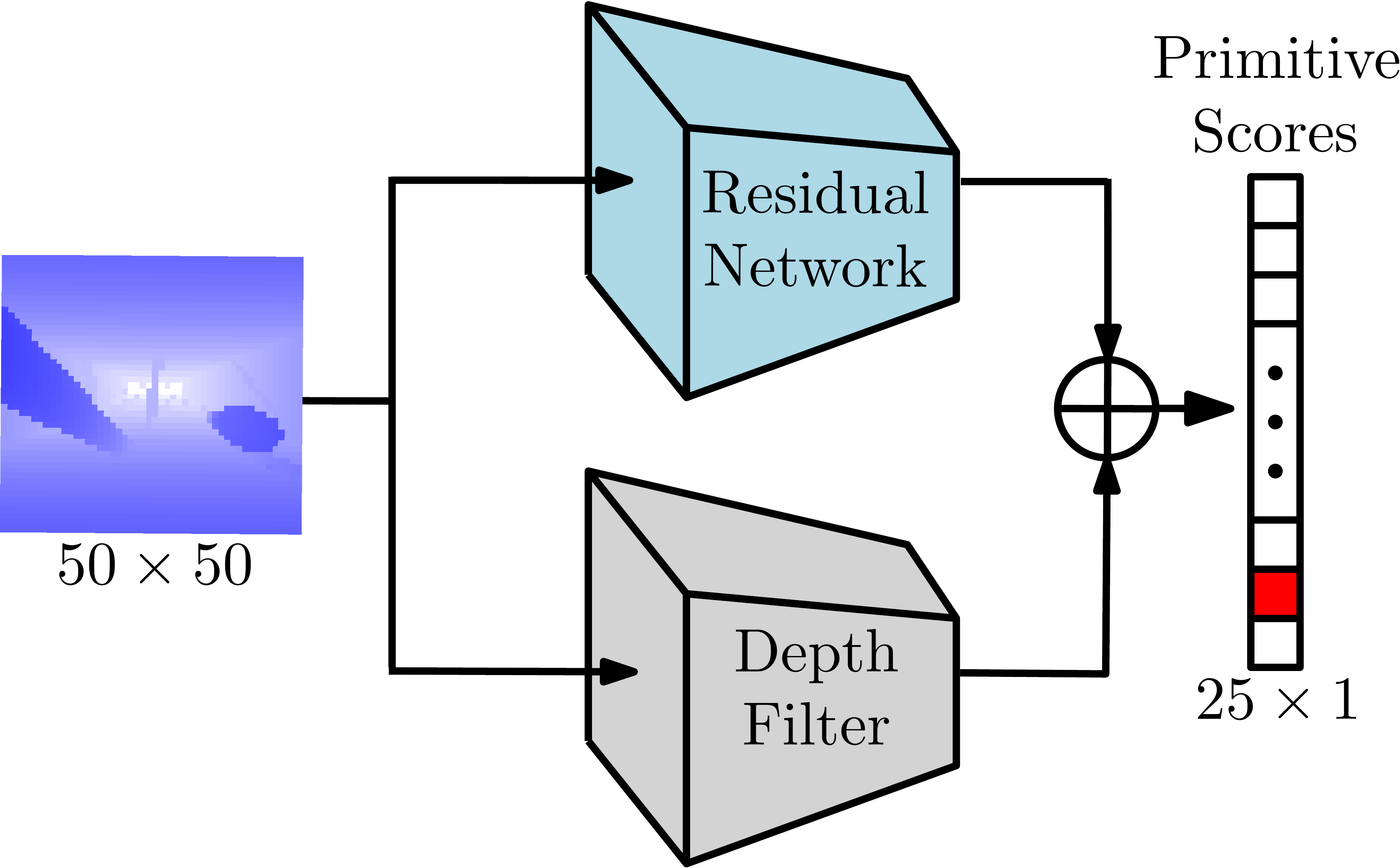}
    \end{center}
    \vspace{-3mm}
    \caption{\small Motion primitive trajectories for the quadrotor. \label{fig:quad-policy}}
    \vspace{-2mm}
\end{wrapfigure}

\begin{figure*}[b]
\centering
\subfigure[]
{
\includegraphics[width=0.20\textwidth]{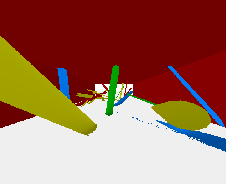}
\label{fig:quadrotor-rgb}
}
\centering
\subfigure[]
{
\includegraphics[width=0.20\textwidth]{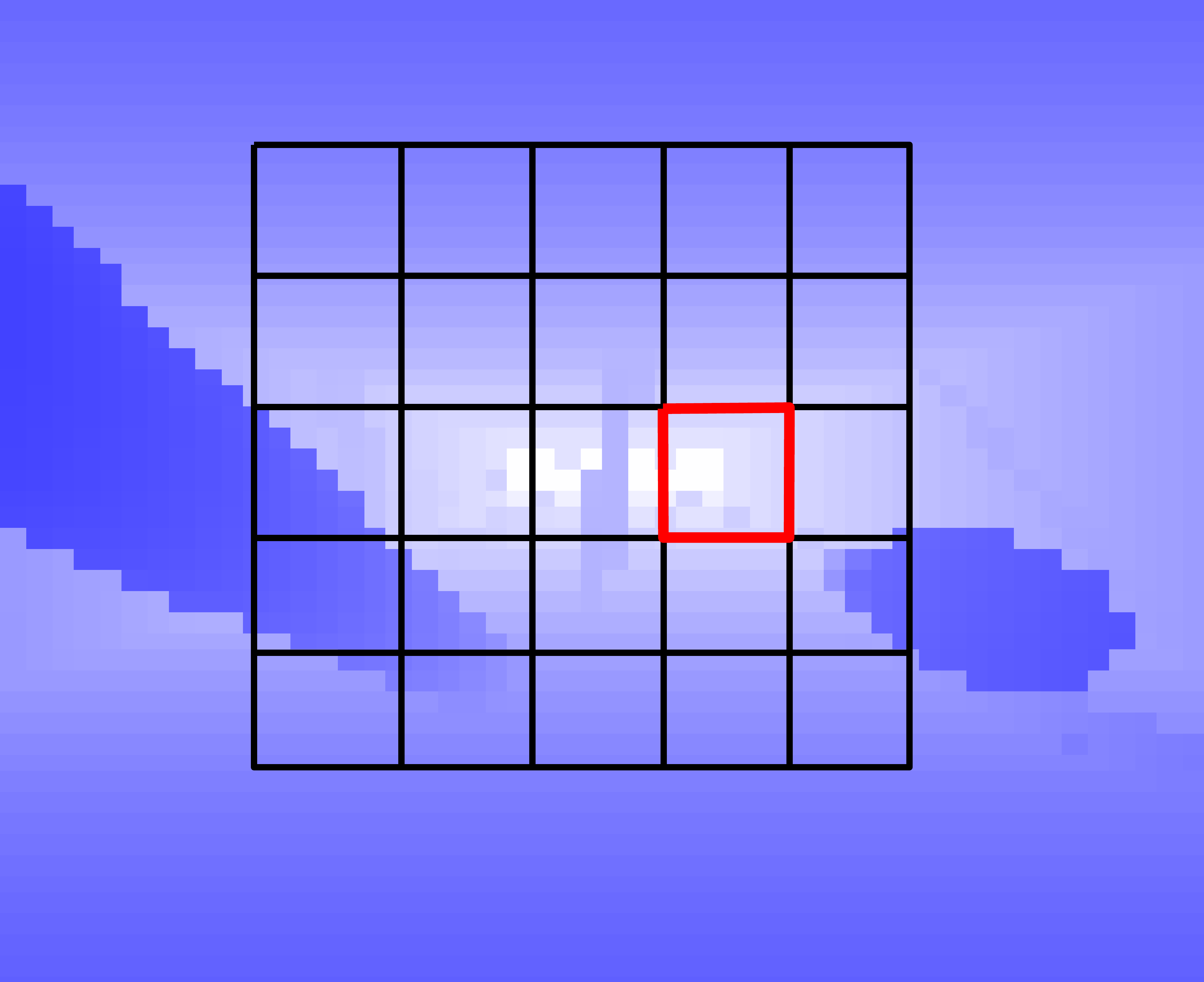}
\label{fig:quadrotor-depth}
}
\subfigure[]
{
\includegraphics[width=0.20\textwidth]{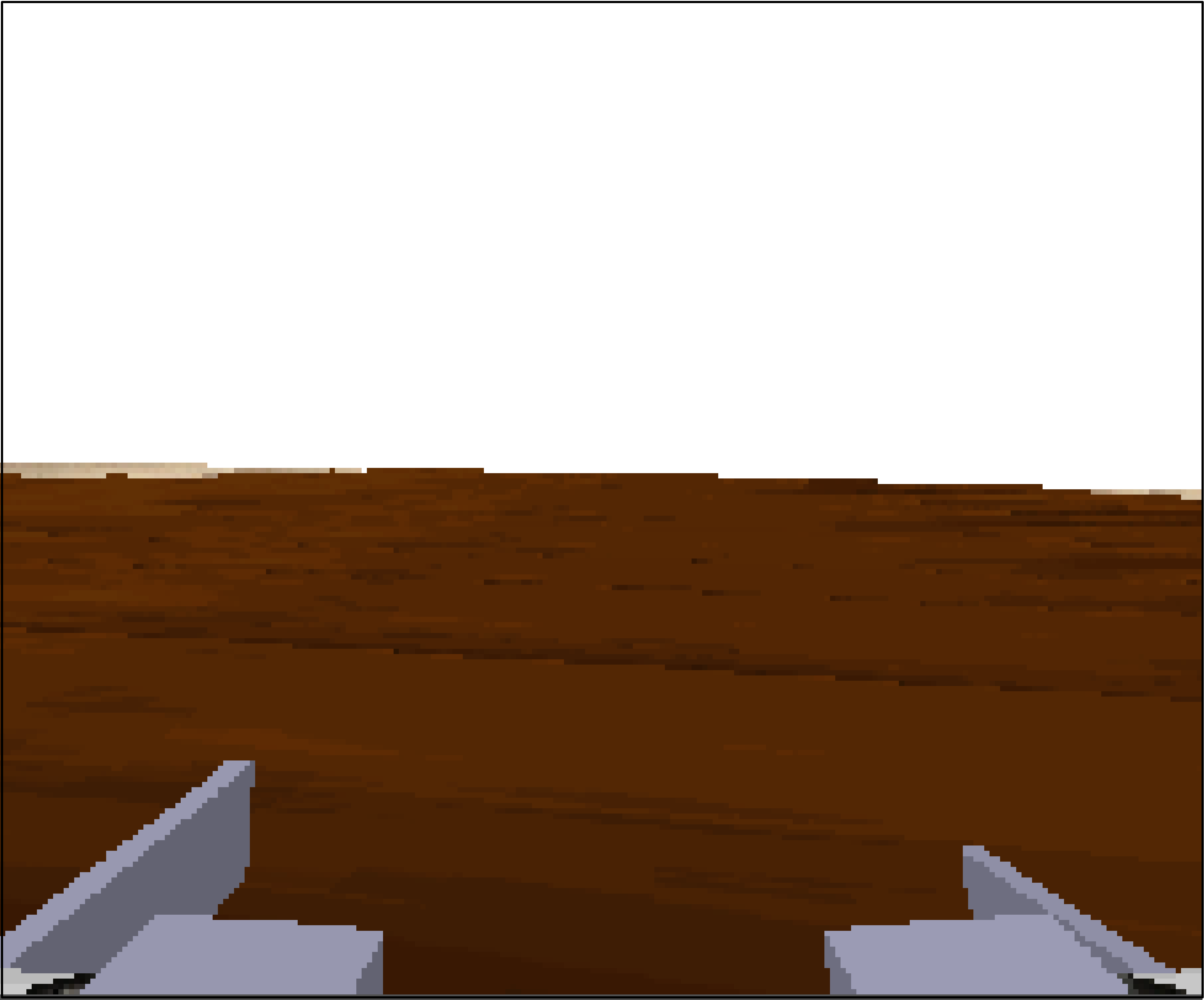}
\label{fig:minitaur-rgb}
}
\centering
\subfigure[]
{
\includegraphics[width=0.20\textwidth]{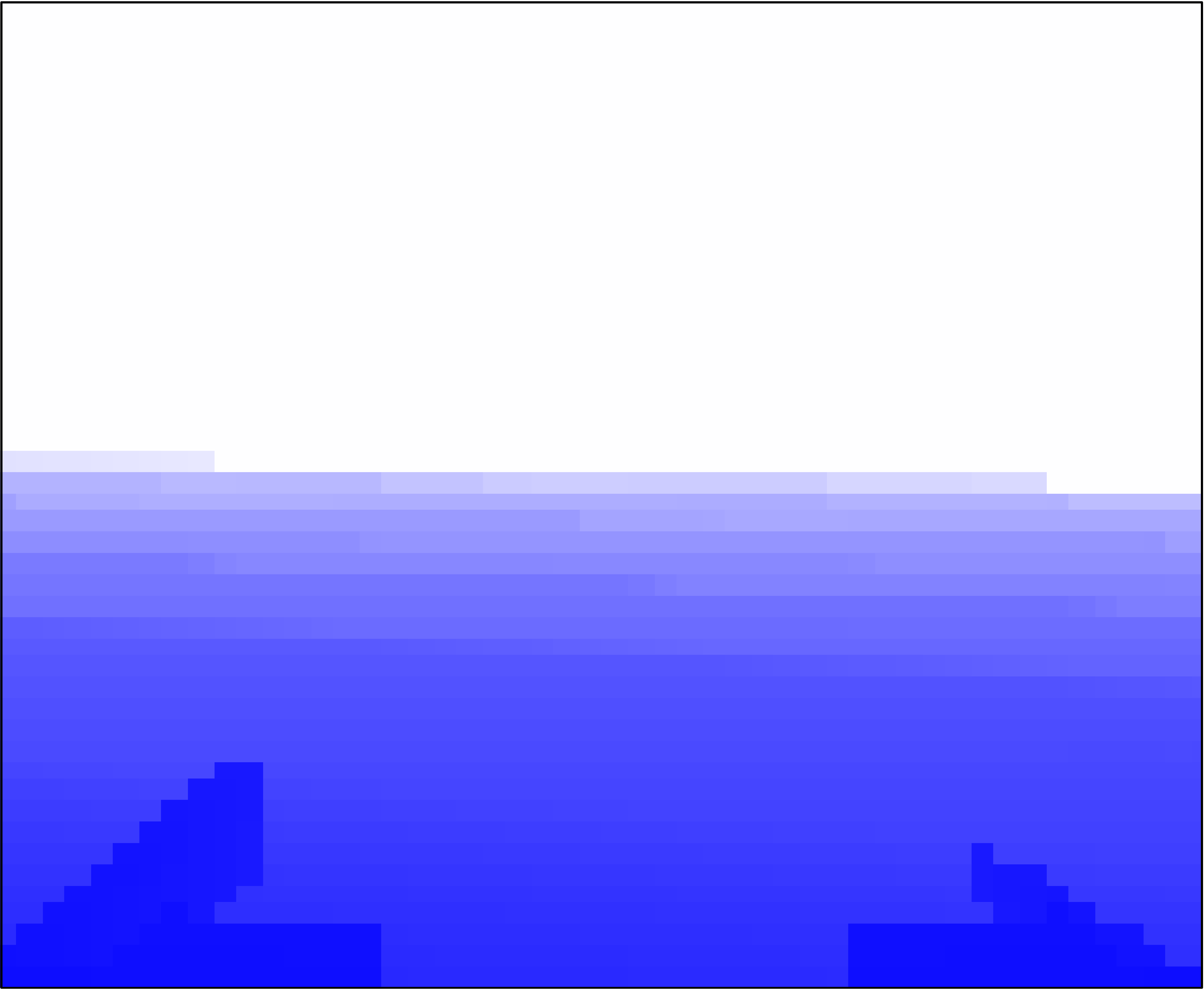}
\label{fig:minitaur-depth}
}
\vskip -10pt
\caption{\small Visualization of the output from the RGB-D camera mounted on the quadrotor and the quadruped. {\bf UAV's onboard camera:} \textbf{(a)} RGB image and \textbf{(b)} Depth map with the Depth Filter grid overlaid on top. The red square highlights the cell withthe highest average depth {\bf Quadruped's onboard camera:} \textbf{(c)} RGB image and \textbf{(d)} Depth map.\label{fig:quad-sensor}
}
\end{figure*}

A visualization of the depth map obtained from the RGB-D camera on the robot can be found in Fig.~\ref{fig:quadrotor-depth} and the corresponding RGB image is shown in Fig.~\ref{fig:quadrotor-rgb} for comparison. 
We model our policy as a DNN with a ResNet-like architecture (illustrated in Fig.~\ref{fig:quad-policy}). 
The policy processes the depth map along two parallel branches: the Depth Filter (which is fixed and has no parameters to learn) and the Residual Network (which is a DNN). Both branches generate a score for each primitive in $\mathcal{L}$ which are summed to obtain the final aggregate score.
The Depth Filter embeds the intuition that the robot can avoid collisions with obstacles by moving towards the ``deepest" part of the depth map. We construct the Depth Filter by projecting the quadrotor's position after executing a primitive $(x_0+\Delta x,y_0+\Delta y,z_0+\Delta z)$ on the depth map to identify the pixels where the quadrotor would end up; see the $5\times 5$ grid in Fig.~\ref{fig:quadrotor-depth} where each cell corresponds to the ending position of a motion primitive. The Depth Filter then applies a mask on the depth map that zeros out all pixels outside this grid and computes the average-depth of each cell in the grid which is treated as a primitive score from that branch of the policy. Note that this score is based only on the ending position of the quadrotor; the entire primitive trajectory when projected onto the depth map can lie outside the grid in Fig.~\ref{fig:quadrotor-depth}. Therefore, to improve the policy's performance, we train the Residual Network. Intuitively, this augments the scores from the Depth Filter branch by processing the \emph{entire} depth map. 
The Residual Network is a DNN with $13943$ parameters: $\texttt{(50,50)} - \texttt{Conv(ELU)} - \texttt{(2,24,24)} - \texttt{Conv(ELU)}-\texttt{(23,23)}-\texttt{Flatten}-\texttt{(529)}-\texttt{Linear(ELU)-(25)-\texttt{Linear(Tanh)-(25)}}$.


\subsection{Quadrupedal Locomotion on Uneven Terrain}
\label{app:minitaur}

A visualization of the depth map used by the policy can be found in Fig.~\ref{fig:minitaur-depth} and the corresponding RGB image is shown in Fig.~\ref{fig:minitaur-rgb} for comparison. The policy we train is a DNN with $944$ parameters: The depth map is passed through a convolutional layer $\texttt{(50,50)} - \texttt{Conv(ELU)} - \texttt{(2,10,10)}-\texttt{Flatten}-\texttt{(200)}$ and the features are concatenated with the $22$ proprioceptive feedback and processed as follows: $\texttt{(222)}-\texttt{Linear(Sigmoid)}-\texttt{(4)}$. Let the 4 outputs be denoted by $\eta_1,\cdots,\eta_4$. Then, the primitive is assigned as: $a_1 = 0.8\eta_1+0.2; ~ a_2 = 0.8\eta_2+0.2; ~ \omega = 20\eta_3 + 20; ~ a_{\rm st} = \max\{\eta_4, \min\{1-a_1,1-a_2\}\}$.
Intuitively, the transformation above from $\eta$ to $[a_1,a_2,\omega,a_{\rm st}]$ ensures that the robot has a minimum forward speed and a maximum steering speed. 

\newpage


\section{Plan for Experiments}
\label{app:expts}

We plan to validate our approach on two hardware examples: UAV navigation across novel obstacle fields, and quadrupedal locomotion on uneven terrain. The necessary hardware for these experiments is readily available in our lab. In both examples, we will train a posterior distribution on the policy space in simulation using PyBullet (exactly as described in the paper) and then transfer it to the robot in a zero-shot manner.

\begin{figure*}[b]
\vspace{-2mm}
\centering
\subfigure[]
{
\includegraphics[height=0.25\textwidth]{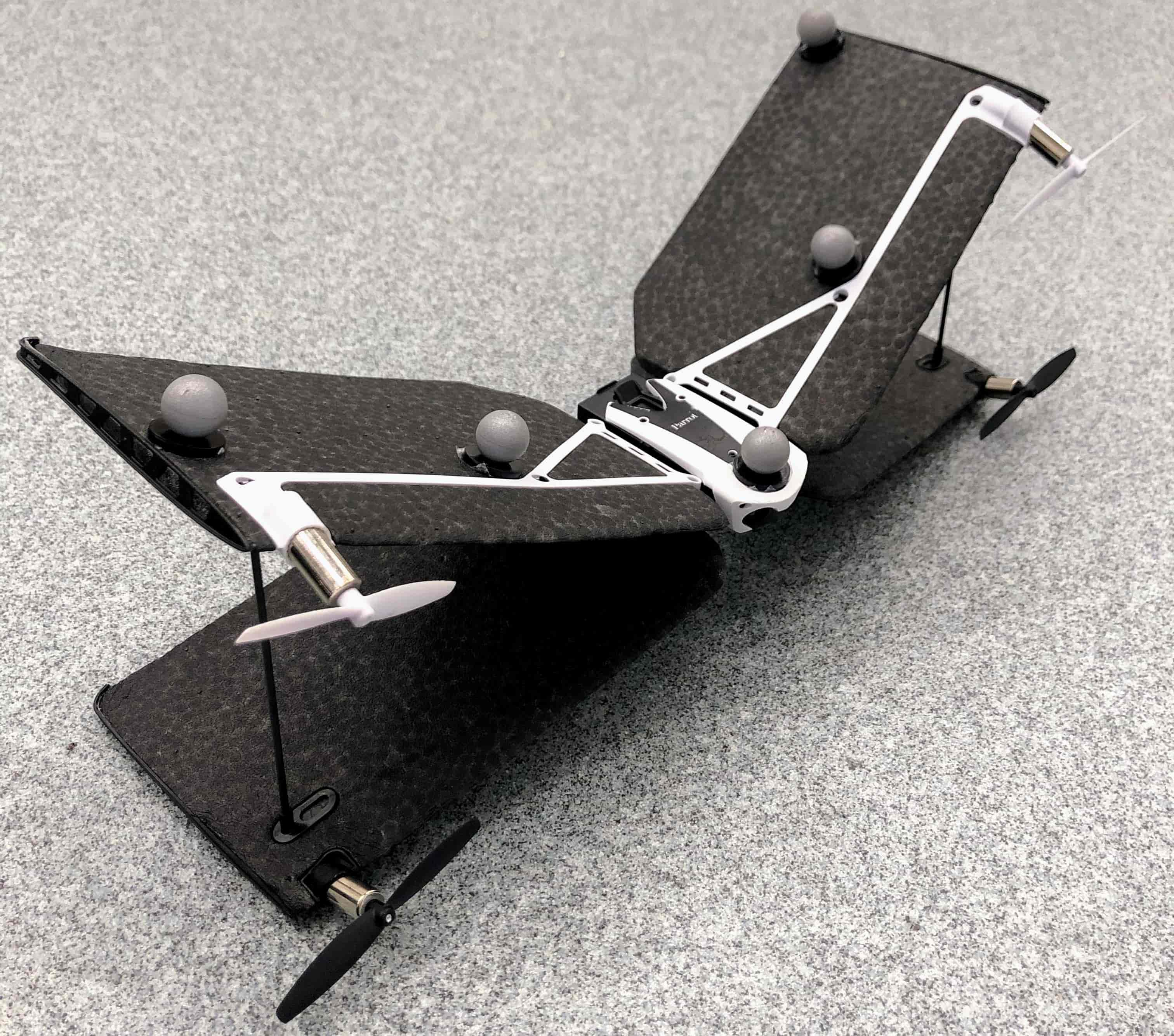}
\label{fig:swing}
}
\centering
\hspace{1mm}
\subfigure[]
{
\includegraphics[height=0.25\textwidth]{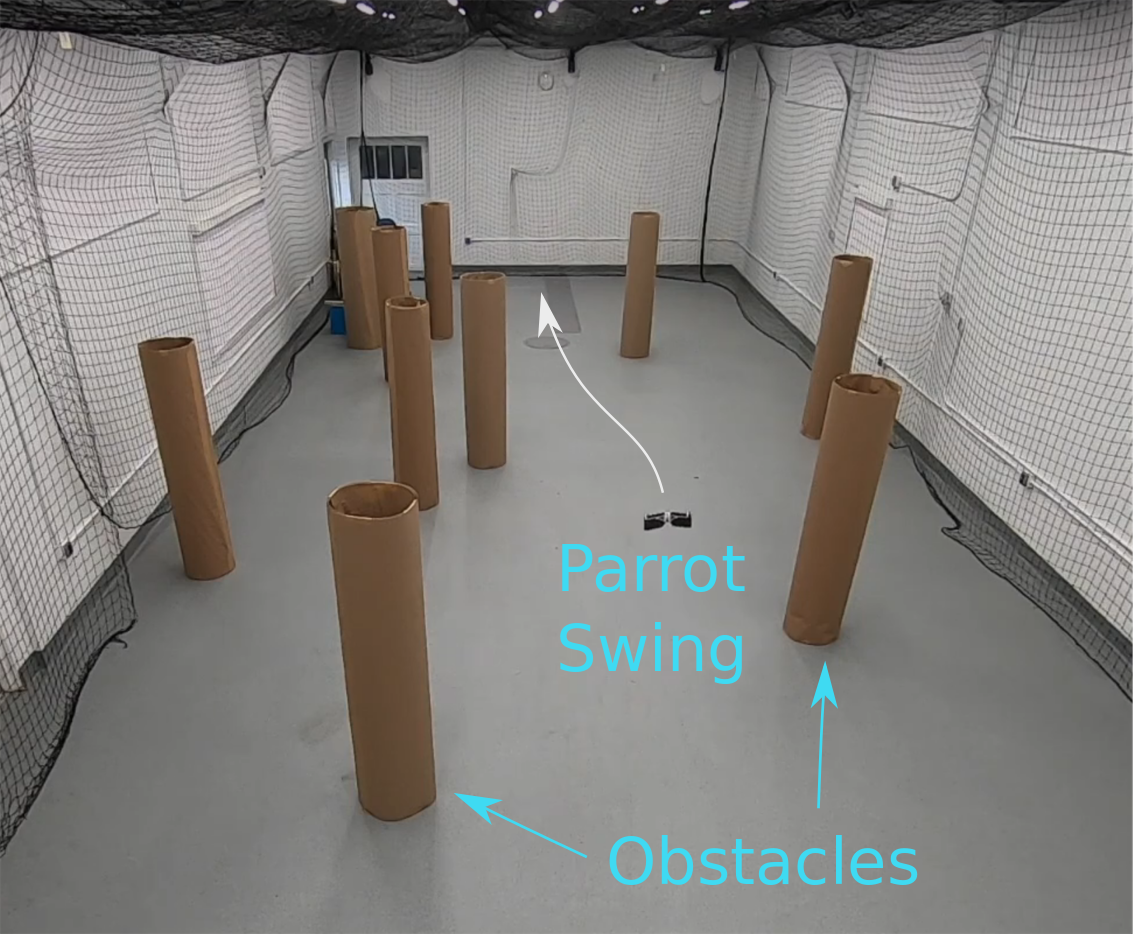}
\label{fig:arena}
}
\hspace{1mm}
\subfigure[]
{
\includegraphics[height=0.25\textwidth]{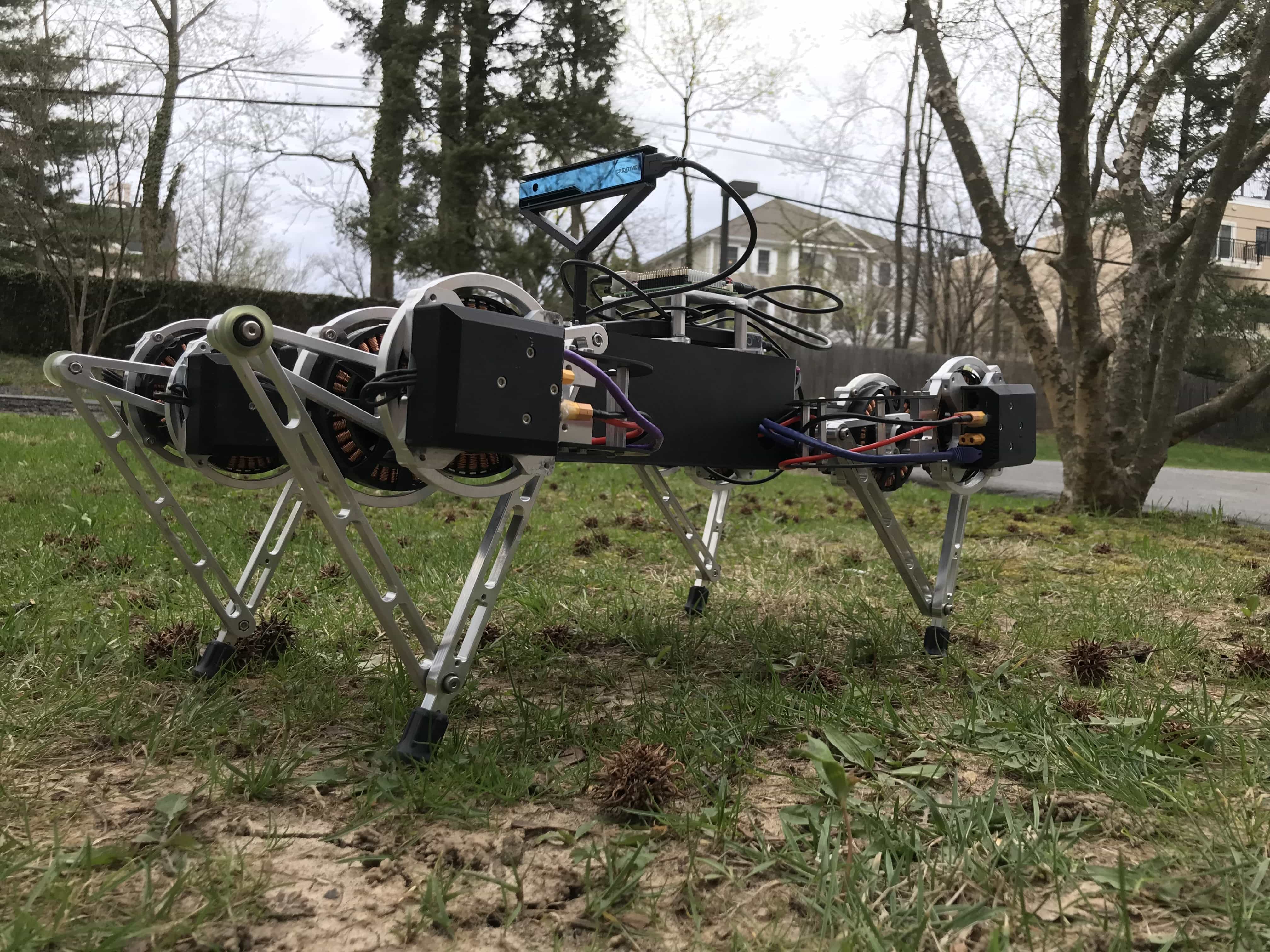}
\label{fig:minitaur-robot}
}
\vskip -10pt
\caption{\small{Experimental setup. \textbf{(a)} Parrot Swing drone with Vicon markers.  \textbf{(b)} An example obstacle course for the UAV. \textbf{(c)} Minitaur mounted with an Intel RealSense RGB-D camera}}
\vspace{-2mm}
\end{figure*}

\vspace{3mm}

\subsection{UAV Navigation}
\vspace{2mm}
{\bf Robot.} We will use the Parrot Swing drone (Fig.~\ref{fig:swing}) for this experiment. This drone is lightweight (75g) and can efficiently travel at speeds greater than 2 m/s by leveraging its ``wings".

{\bf Motion Primitives.} To generate the library of motion primitives $\mathcal{L}$, we will first compute the primitives -- as described in Section~\ref{subsec:quadrotor} -- using the robot's dynamical model. We will then execute each primitive on the hardware system and the resulting trajectories will be recorded by a motion capture system (Vicon). These hardware-generated trajectories will form our library for the training in simulation.

{\bf Experimental Setup.} Our obstacle course will have randomly placed cardboard cylinders in a 7m~$\times$~18m netted area with Vicon; see Fig.~\ref{fig:arena}. The drone will fly from one end of this netted area to another and the cost will be assigned based on the percentage of the obstacle course it navigates without collision. The depth map for the policy will be synthetically generated using Vicon. We expect that the sim-to-real gap will be tractable in our experimental setup because: (i) depth maps, unlike RGB images, are independent of the texture of the scene, and (ii) the true motion primitive trajectories recorded with Vicon will mitigate system identification errors.

\vspace{3mm}

\subsection{Quadrupedal Locomotion}
\vspace{2mm}
{\bf Robot.} We will use the quadrupedal robot Minitaur \cite{Kenneally16} (Fig.~\ref{fig:minitaur-robot}) for this experiment. We have mounted an Intel RealSense RGB-D camera on the torso of the robot (as shown in Fig.~\ref{fig:minitaur-robot}) for feedback of the forthcoming terrain's depth map.

{\bf Motion Primitives.} We use a library of ``\texttt{sine}" controllers which are provided with Minitaur's SDK, developed by Ghost Robotics, and is readily available for implementation on hardware; see Section~\ref{subsec:minitaur} for further details on the motion primitives.

{\bf Experimental Setup.} We will construct uneven terrains by connecting wood planks with hinges that will allow us to randomize the step-to-step angle of the rough terrain (analogous to the terrain shown in Fig.~\ref{fig:minitaur}). The cost of a rollout will be determined by the percentage of the terrain the robot traverses successfully. We expect the sim-to-real gap to be tractable because (i) we use the original-equipment-manufacturer's (OEM) controllers in the library of motion primitives; (ii) PyBullet's Minitaur API, which is used for training the posterior distribution, models the robot in great detail, including features such as motor shut-off due to overheating; and (iii) we use depth maps for terrain feedback instead of RGB images which are independent of the scene's texture.

\end{document}